\documentclass[conference]{IEEEtran}
\usepackage[pdftex]{graphicx}
\usepackage{amsfonts}
\usepackage[cmex10]{amsmath}
\usepackage{mathabx}
\usepackage{algorithmic}
\usepackage{array}
\usepackage{mdwmath}
\usepackage{mdwtab}
\usepackage{eqparbox}
\usepackage{url}
\usepackage{subcaption}
\usepackage{float}
\usepackage{booktabs,chemformula}
\usepackage{hyperref}
\usepackage{caption}
\usepackage{cleveref}
\usepackage{fancyhdr}
\usepackage[super]{nth}
\usepackage[pscoord]{eso-pic}
\usepackage{amssymb}
\makeatletter

\newcommand\AtPageUpperMyright[1]{\AtPageUpperLeft{
     \put(\LenToUnit{0.5\paperwidth},\LenToUnit{-1cm}){
         \parbox{0.5\textwidth}{\raggedleft\fontsize{9}{11}\selectfont #1}}
     }}
\newcommand{\confheader}[1]{
    \AddToShipoutPictureBG*{
    \AtPageUpperMyright{#1}
    }
}
\def\endthebibliography{%
	\def\@noitemerr{\@latex@warning{Empty `thebibliography' environment}}%
	\endlist
}
\makeatother
\begin{document}
    \title{Bengali Handwritten Digit Recognition using CNN with Explainable AI}
    \author{\authorblockN{
        Md Tanvir Rouf Shawon\authorrefmark{1} 
        \textsuperscript{\textsection}, 
        Raihan Tanvir\authorrefmark{2} 
        \textsuperscript{\textsection},
        Md. Golam Rabiul Alam\authorrefmark{3}}
        \authorblockA{\authorrefmark{1}Ahsanullah University of Science and Technology, Dhaka, Bangladesh}
        \authorblockA{\authorrefmark{2}\authorrefmark{3}Brac University, Dhaka, Bangladesh}
        \authorblockA{\{\authorrefmark{1}shawontanvir95, \authorrefmark{2}raihantanvir.96\}@gmail.com, \authorrefmark{3}rabiul.alam@bracu.ac.bd}}
        \maketitle
	\begingroup\renewcommand\thefootnote{\textsection}
            \footnotetext{MD Tanvir Rouf Shawon and Raihan Tanvir have equal contributions.}
        \endgroup
\begin{abstract}
Handwritten character recognition is a hot topic for research nowadays. If we can convert a handwritten piece of paper into a text-searchable document using the Optical Character Recognition (OCR) technique, we can easily understand the content and do not need to read the handwritten document. OCR in the English language is very common, but in the Bengali language, it is very hard to find a good quality OCR application. If we can merge machine learning and deep learning with OCR, it could be a huge contribution to this field. Various researchers have proposed a number of strategies for recognizing Bengali handwritten characters. A lot of ML algorithms and deep neural networks were used in their work, but the explanations of their models are not available. In our work, we have used various machine learning algorithms and CNN to recognize handwritten Bengali digits. We have got acceptable accuracy from some ML models, and CNN has given us great testing accuracy. Grad-CAM was used as an XAI method on our CNN model, which gave us insights into the model and helped us detect the origin of interest for recognizing a digit from an image.
\end{abstract}
	
	
\section{Introduction}
The term "OCR" stands for "Optical Character Recognition." It is a technique for detecting text within a digital image. Text recognition in scanned documents and photographs is a typical application. OCR software can convert a physical paper document or an image into an electronic text-searchable version. There are many programs that can easily convert a digital image into an editable document. Some of the work, like OCR from printed paper using the RNN network ~\cite{parthiban2020optical} and the open-vocabulary OCR system ~\cite{cai2017open}, has great accuracy in character recognition in the English language. There is also a lot of work on translating the handwritten papers into documents. Most of them are in the English language. A novel work ~\cite{8858545} by Rumman Rashid Chowdhury et al. on Bengali handwritten character recognition combining CNN with augmentation for $50$ Bengali characters can be referred to here where they have achieved $95.25\%$ accuracy. If we can combine machine learning approaches to recognize handwritten characters, then it will be a great contribution to the Bengali OCR arena.

As previously stated, much work has been done in this field, but it is not complete because they are relying solely on the model. Explainable AI models are largely used nowadays to explain models that were supposed to be black boxes for us previously. A lot of XAI methods are used to understand normal feature-based datasets where we can measure the participation of any particular feature in a model to predict something. But for images, it is quite hard to visualize the participation of the pixels in prediction. Grad-CAM is the solution to this problem. In our research, we employed Grad-CAM to show the regions that are responsible for the final forecast. our contribution in this paper can be summarized into the following points -
\begin{itemize}
    \item We have performed a handwritten digit recognition task on two widely used datasets, \textit{Ekush} and \textit{NumtaDB}, which contain a large number of digit and character images.
    \item In the search for a better model for character recognition, several machine learning algorithms, and a Convolutional Neural Network has been employed.
    \item We have used the Grad-CAM class activation map as an Explainable AI tool for our task to explain the CNN model we have trained.
\end{itemize}


	\section{Related Works}
	
	Earlier recognition of handwriting was done using the different image-processing techniques. A work by T. K. Bhowmik et al. on handwritten character recognition using basic stroke features \cite{bhowmik2004recognition} has shown us an earlier approach to recognizing handwritten characters where they have used multi-layer perceptron (MLP) as a classifier. A work on English handwritten character recognition \cite{pal2010handwritten} by Anita Pal et al. can also be shown as an earlier work. They also used MLP but they took skeletonization, normalization,  binarization, and most importantly Fourier transform into consideration. The paper on handwritten digit recognition by Cheng-LinLiu et al. used 3 English handwritten digit datasets which are CENPARMI, CEDAR, and MNIST, and applied different state-of-the-art models like K nearest neighbor, quadratic discriminant function, and SVC on them. They have claimed that the accuracy they have got is quite competitive to the best ones previously reported on the same datasets. The paper BornoNet ~\cite{RABBY2018528} by Akm Shahariar AzadRabby et al. used CNN for classification on three different datasets which are BanglaLekha-Isolated dataset, ISI and CMATERdb. The validation score they found is $95.71\%$, $96.81\%$ $98\%$. MM Rahman et al. also applied CNN on their prepared dataset containing 20000 images having 400 samples for each character on their paper \cite{rahman2015bangla} and got accuracy of $93.93\%$ and $85.36\%$ for training and test sets respectively. The work by RR Chowdhury et al. used data augmentation on the dataset BanglaLekhaIsolated \cite{BISWAS2017103} and applied CNN to it. They have found an accuracy of $94.576\%$ and a loss of $0.204$.
	
A review paper by Tapotosh Ghosh et al. has reviewed various works currently available on Bengali handwritten character recognition and compared the performance of their work. The authors have nicely listed the available papers from which we can easily understand the impact of machine learning and deep learning on Bengali handwritten character recognition.

	\section{Dataset}
    We have used a dataset which is called \textit{Ekush}. This dataset is quite similar to the \textit{MNIST} \cite{cohen2017emnist} dataset. The dimension of the images of both datasets is $28*28$. It is the largest dataset for character recognition, having handwritten Bangla characters \cite{10.1007/978-981-13-9187-3_14}. The dataset consists of Bengali vowels, modifiers, consonants, compounds, and numerical digits. It has a total of $367,018$ handwritten characters which are isolated and written by $3086$ individual writers. The dataset is also divided into gender and age groups. There is an equal number of data for males and females in this dataset. As we are using only the digits, we have taken the digit dataset for both males and females. Here, the male dataset has $15,208$ digit and the female dataset has $15,622$ digit. So we have used $30,830$ digits in total. Table ~\ref{tab:dataset-frequency} contains the frequency of individual digits. We can see that the dataset is almost balanced. So we did not modify the images.
    
	\begin{table}[H]
		\centering
		\begin{tabular}{|c|c|c|c|}
			\toprule
			Digit  &	Size  & Digit & Size\\
			\midrule
			$0$ & $3083$ & $5$ & $3081$ \\
        $1$ & $3084$ & $6$ & $3083$ \\
        $2$ & $3075$ & $7$ & $3086$ \\
        $3$ & $3090$ & $8$ & $3087$ \\
        $4$ & $3087$ & $9$ & $3074$ \\
			\bottomrule
		\end{tabular}
		\caption{Frequency of individual digits in \textit{Ekush} dataset}
		\label{tab:dataset-frequency}
	\end{table}
	
\begin{figure}
		\centering
		\begin{subfigure}[b]{.22\columnwidth}
		    \centering
			\includegraphics[width=1\linewidth]{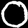}
		\end{subfigure}
		\begin{subfigure}[b]{.22\columnwidth}
		    \centering
			\includegraphics[width=1\linewidth]{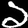}
		\end{subfigure}
		\begin{subfigure}[b]{.22\columnwidth}
		    \centering
			\includegraphics[width=1\linewidth]{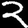}
		\end{subfigure}
		\begin{subfigure}[b]{.22\columnwidth}
		    \centering
			\includegraphics[width=1\linewidth]{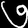}
		\end{subfigure}
		\par\medskip
		\begin{subfigure}[b]{.22\columnwidth}
		    \centering
			\includegraphics[width=1\linewidth]{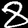}
		\end{subfigure}
		\begin{subfigure}[b]{.22\columnwidth}
		    \centering
			\includegraphics[width=1\linewidth]{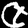}
		\end{subfigure}
		\begin{subfigure}[b]{.22\columnwidth}
		    \centering
			\includegraphics[width=1\linewidth]{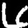}
		\end{subfigure}
		\begin{subfigure}[b]{.22\columnwidth}
		    \centering
			\includegraphics[width=1\linewidth]{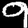}
		\end{subfigure}
		\caption{Some sample images from \textit{Ekush} dataset. Each column containing 4 images having the digits from $0 - 3$ and $4 - 7$ respectively.} 
		\label{fig:ekush_image}
	\end{figure}

	We have got the dataset from Kaggle as an individual spreadsheet. We found that the spreadsheets have $785$ columns, of which the last one was the label. They have labeled the digits from $112$ to $121$ representing $0$ to $9$. At first, we took the individual male and female datasets in a data-frame and dropped the label. The data set contains $28 \times 28$ pixel grayscale images, and they are already flattened. So we did not need to flatten them again. Then we merged the male and female datasets together as our final data. The labels were also taken to another array as the target. We subtracted $112$ from all the labels and made them $0$ to $9$. Finally, all the flattened images were normalized by $255$ as the highest value of a grayscale image is $255$. So now we have a normalized dataset of size $30,830$ having values from $0 - 1$. We can see some sample images from the \textit{Ekush} dataset in Fig. ~\ref{fig:ekush_image}.
	
\begin{figure}
		\centering
		\begin{subfigure}[b]{.3\columnwidth}
		    \centering
			\includegraphics[width=1\linewidth]{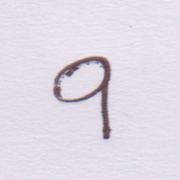}
		\end{subfigure}
		\begin{subfigure}[b]{.3\columnwidth}
		    \centering
			\includegraphics[width=1\linewidth]{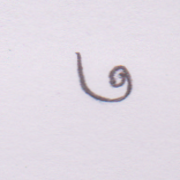}
		\end{subfigure}
		\begin{subfigure}[b]{.3\columnwidth}
		    \centering
			\includegraphics[width=1\linewidth]{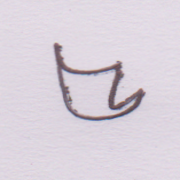}
		\end{subfigure}
		
		\setlength{\belowcaptionskip}{-18pt}
		\caption{Some sample images from \textit{NumtaDb} dataset.}
		\label{fig:namta_image}
	\end{figure}

    We partially used another dataset called \textit{NumtaDB}\footnote{https://www.kaggle.com/BengaliAI/numta} which was also taken from Kaggle containing $72,045$ unique images of Bengali digits, from which we took $54908$ images for the experiment. The processing of this dataset was also done in the previously described ways, but here the images are in RGB format. So we had to convert the images to grayscale. We also resized the images to $28*28$ pixels. Some samples are shown in Fig. ~\ref{fig:namta_image}.	

	\section{Methodology and Experiments}
	
	Machine learning algorithms are mostly used nowadays for any kind of classification. We must apply some of the basic machine learning  algorithms because we are primarily working with images classification. Some of them are described here.  
	\subsection{Machine Learning Models}
	First, we used \textbf{decision tree classifier}. A decision tree is a flowchart-like structure in which each internal node represents a feature or characteristic, each branch indicates a criterion, and each leaf reflects the outcome. In a decision tree, the source node is the uppermost node. We used GINI impurity for the calculation and made a decision based on the calculation.\\
Next, \textbf{random forest classifier} is an ensemble approach of decision trees built on a randomly partitioned database (based on the divide-and-conquer methodology). A cluster of decision trees is referred to as a "forest." Information gain, GINI index, or gain ratio, which is attribute selection indicators, create the individual decision trees. An independent random sample creates each of the trees that make up a forest.\\
	\textbf{K nearest neighbor} is another algorithm that calculates the distances of the testing sample from every train data point and decides which class it should go into. We used $10$ as our parameter in the K nearest neighbor algorithm.\\
We also used \textbf{Support vector machine (SVM)} as our classifier. Here the decision boundary updates its' weight using the train data and classifies the test ones in SVM. \\
We also used \textbf{NuSVM} which is similar to the SVM in terms of mathematical calculation. But there is a difference between them in parameters. SVM uses C as a regularization parameter, whereas NuSVC uses nu, which is an upper bound on the fragment of marginal miscalculation and a lower bound on the fragment of support vectors. It should be in the interval of $(0, 1)$.\\
\textbf{AdaBoost} is an aggregation strategy that combines a group of weak trainees to develop a solid learner. Usually, decision stumps created by each weak learner are used to classify the observations.\\
\textbf{GradientBoosting} is similar to AdaBoost but works with residuals to build an additive model. It also introduces the learning rate.\\
\textbf{Naive Bayes} is based on Bayes' formula, but it has a very naive assumption, which is the assumption of independence. We have used Gaussian Naive Bayes as our data is continuous. Here is the equation ~\ref{fig:naive} for the naive Bayes algorithm.
	\begin{equation}
	    P(c/X)=P(x_{1}/c)*P(x_{2}/c)*...*P(x_{n}/c)*P(c)
	    \label{fig:naive}
	\end{equation}
	
    \textbf{Linear discriminant and quadratic discriminant} are two algorithms where a line and a curve are used to classify the test data, respectively.
\subsection{Convolutional Neural Network}
We applied a \textbf{convolutional neural network} simultaneously on the \textit{Ekush} dataset. CNN and artificial neural networks work pretty similarly. A simple structure in an image can be detected through convolution. CNN is basically used to recognize objects in an image. It uses a feed-forward neural network to classify any complex structure in a digital image. The three layers it uses are input, hidden, and output layers. A kernel of a specific size traverse through the image and helps to detect a particular pattern.

	\section{Experimental Result}
    \subsection{Experimental Setup:}
    Both the \textit{Ekush} and \textit{NumtaDB} datasets were run through the previously mentioned machine learning algorithms. 
In both experiments, datasets were split into train and test sets containing $80\%$ and $20\%$ of the images respectively.  
We used the sklearn package to use the algorithms. For Grad-CAM we used a combination of Python $3.6.5$, Keras $2.3.1$, Tensorflow $2.0.0$ and Keras-vis. Google Colaboratory was used to implement our codes.\\	
    
    For both datasets, the parameters were the same for every algorithm. For K Nearest Neighbour K was $10$, SVC classifier had kernel="RBF" which is radial basis function kernel, $C=0.025$, probability=True. In decision tree and random forest classifiers, the maximum depth was not defined. So it was set to "none", as the default value which means that it will expand the nodes until all the leaves of that tree are pure. All the other algorithms are kept as they are in their default function in sklearn. We have used \textbf{Accuracy} as a metric to judge the models and also calculated the log loss, which is the average of the sum of the log of improved forecast probabilities for each data point, which can be defined using the Eq. ~\ref{eq:logloss}. It is better when the log loss is lower. It means the prediction is better.
	\begin{equation}
	    -\frac{1}{N}\sum_{k=1}^Ny_{k}.\log{(p(y_{k}))}+(1-y_{k}).\log{(1-p(y_{k}))}
	    \label{eq:logloss}
	\end{equation}
 The comparison of accuracy and log loss of different models has been shown using bar charts. For visualizing the comparison, we have used the seaborn and matplotlib packages of Python.
\subsection{Result Analysis}
Accuracy and log loss for the 10 algorithms we have used are given in Table ~\ref{tab:ekushe_table} for \textit{Ekush} and \textit{NumtaDB} dataset for a better understanding of the result.

\begin{figure}
		\centering
			\includegraphics[width=1\linewidth]{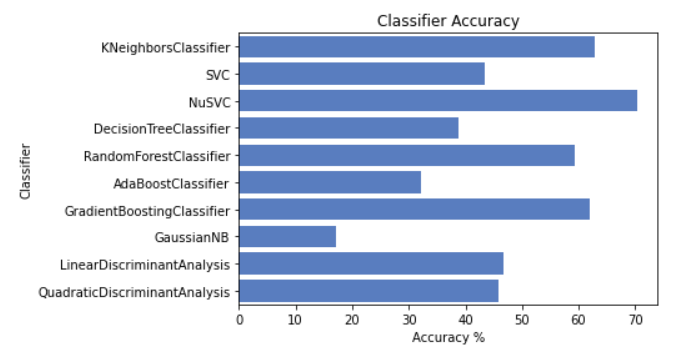}
		\caption{Comparison of accuracy among the used algorithms for \textit{NumtaDB} dataset.}
		\label{fig:namta_comparison}
	\end{figure}

\begin{figure}
		\centering
			\includegraphics[width=1\linewidth]{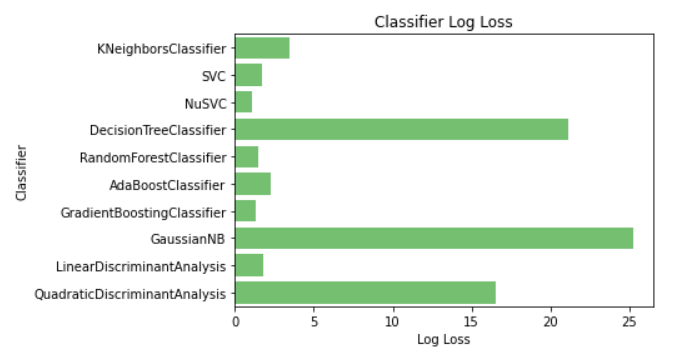}
		\caption{Comparison of log loss among the used algorithms for \textit{NumtaDB} dataset.}
		\label{fig:namta_log}
	\end{figure}

It can be seen that \textbf{NuSVC} algorithm has the highest accuracy of \textbf{$70.445108\%$} and log loss of \textbf{$1.063555$} owning the lowest of all the algorithms which can be seen in Fig ~~\ref{fig:namta_comparison} and ~\ref{fig:namta_log} respectively for \textit{NumtaDB} dataset. So it is evident that NuSVC is the best classifier for the \textit{NumtaDB} dataset. As the highest accuracy for \textit{NumtaDB} is only $70.45\%$, we did not explore it anymore.
	
\begin{figure}
		\centering
			\includegraphics[width=1\linewidth]{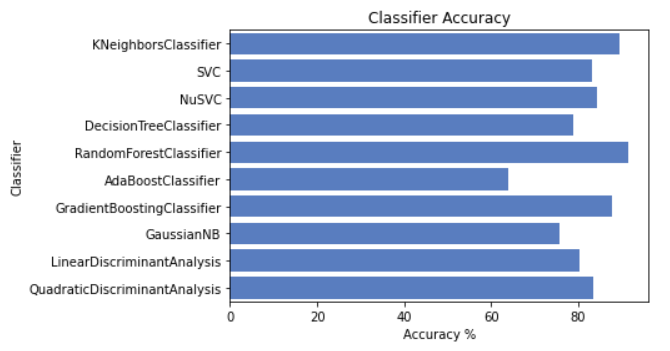}
            
		\caption{Comparison of accuracy among the used algorithms for \textit{Ekush} dataset.}
		\label{fig:ekush_comparison}
	\end{figure}

\begin{figure}
		\centering
			\includegraphics[width=1\linewidth]{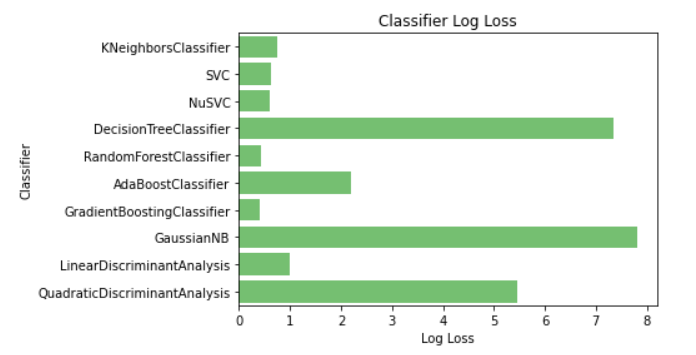}
            \setlength{\belowcaptionskip}{-14pt}
		\caption{Comparison of log loss among the used algorithms for \textit{Ekush} dataset.}
		\label{fig:ekush_log}
	\end{figure}

\begin{figure}
		\centering
			\includegraphics[width=1\linewidth]{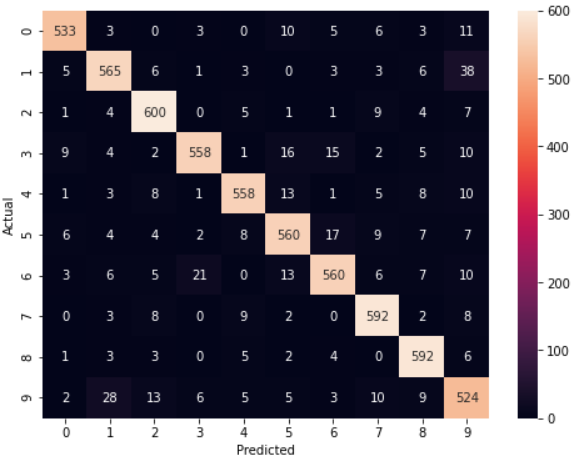}
		\caption{Confusion matrix of the Random Forest Classifier model trained on \textit{Ekush} dataset.}
		\label{fig:confusion}
	\end{figure}

We can see that \textbf{Random Forest} classifier has the highest accuracy of \textbf{$91.5018\%$} and log loss of \textbf{$0.442208$} owning the second lowest of all the algorithms which can be seen in Fig ~\ref{fig:ekush_comparison} and ~\ref{fig:ekush_log} respectively for \textit{Ekush} dataset. Here we have also shown the confusion matrix for the random forest classifier in Fig ~\ref{fig:confusion} where we can notice that the ratio of correct predictions is quite high than the incorrect ones and an interesting observation is the higher number of the wrong prediction of the digit $1$ with the digit $9$ and vice versa as both the digit seems very close when written in bare hands. After observing the results, it is clear to us that Random Forest is the best classifier among the machine learning algorithms for the \textit{Ekush} dataset.

	\begin{table*}[]
		\centering
		
		\begin{tabular}{|c|c|c|c|c|}
			\toprule
            \textbf{Classifier} &  \textbf{Accuracy(\%)} & \textbf{Log Loss} & \textbf{Accuracy(\%)} & \textbf{Log Loss}\\  
			\midrule
			KNeighborsClassifier & $89.312400$ & $0.758313$ & $62.890653$ &  $3.438639$\\
                            SVC  & $83.279273$ & $0.620342$ & $43.490930$ &  $1.735782$\\
                          NuSVC  & $84.187480$ & $0.613189$ & \textbf{70.445108} &  \textbf{1.063555}\\
         DecisionTreeClassifier  & $78.738242$ & $7.343551$ & $38.675603$ & $21.180696$\\
         RandomForestClassifier  & \textbf{91.501800} & \textbf{0.442208} & $59.248197$ &  $1.447900$\\
             AdaBoostClassifier  & $63.833900$ & $2.198822$  & $32.162891$ &  $2.270909$\\
     GradientBoostingClassifier  & $87.609500$ & $0.411749$ & $62.052888$ &  $1.305641$\\
                     GaussianNB  & $75.559500$ & $7.821966$ & $17.214249$ & $25.294058$\\
     LinearDiscriminantAnalysis  & $80.327600$ & $0.986408$ & $46.638013$ &  $1.764705$\\
  QuadraticDiscriminantAnalysis  & $83.376600$ & $5.471474$ & $45.880382$ & $16.527047$\\
  Convolutional Neural Network & \textbf{96.707749} & \textbf{0.132881}  & $-$ & $-$\\
			\bottomrule
		\end{tabular}
		\caption{Accuracy and Log Loss of different machine learning algorithms for \textit{Ekush} in coloumn $2-3$ and \textit{NumtaDB} dataset in coloumn $4-5$.}
		\label{tab:ekushe_table}
	\end{table*}

\begin{figure}
		\centering
			\includegraphics[width=1\linewidth]{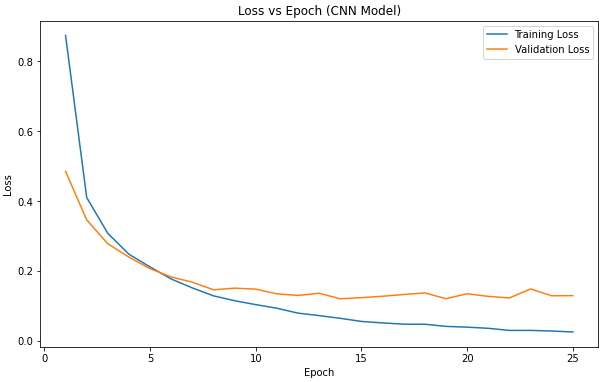}
            \setlength{\belowcaptionskip}{-10pt}
		\caption{Training and Validation loss for CNN model with respect to the Epoch.}
		\label{fig:cnn_graph}
	\end{figure}

As we have achieved a satisfactory result from \textit{Ekush} dataset, we decided to explore it more. Next, we applied a convolutional neural network (CNN) on this dataset. The configuration of our CNN model -- batch size was set to $250$, epoch was $25$. We used the sequential model Keras package for implementing CNN. There were 3 hidden layers in our CNN model. The first hidden layer had $32$ neurons, the kernel size was $(3,3)$ and the activation function was a rectified linear unit (ReLU). Then the second hidden layer had $64$ neurons, the kernel size was $(3,3)$, and the activation function was again a rectified linear unit (ReLU). Then the third hidden layer had $256$ neurons, which is a dense layer where all the inputs are flattened and the activation function was ReLU. Finally, a softmax-activated layer is used for performing multiclass classification at the output layer. \textbf{Adam} was the optimizer of the model and categorical cross-entropy loss was used as a loss function. We also used max-pooling of size $(2,2)$. Again dropout of $25\%$ neuron was done randomly to avoid overfitting in the model. The dataset was split into three sections, which are the train set, validation set, and test set, having a ratio of $60\%$,$20\%$, and $20\%$ respectively. We also renamed the last layer, as we needed it for Grad CAM visualization. After $25$ epoch we got the training accuracy of $99.20\%$ and loss of $0.0251$, validation accuracy of \textbf{96.39\%}, and loss of \textbf{0.1291}. The progress of log loss of both the training and validation sets in the model is shown here in Fig ~\ref{fig:cnn_graph}. And lastly, the test accuracy of $96.71\%$ and loss of $0.1329$ is very good accuracy and better than the accuracy of our previously experimented random forest model. So, using CNN, we have achieved great accuracy in the recognition of Bengali handwritten digits.

	\section{GRAD-CAM CLASS ACTIVATION MAPS}
The invisible layers of a neural network are black boxes to us. If we want to know the hidden layers of a neural network, we can use different \textbf{Explainable AI} methods. There are different XAI methods for explaining a trained model, like \textbf{LIME}, \textbf{SHAP}, \textbf{CAM}. In describing a neural network model where gradients are used Grad-CAM is the perfect candidate. As we are working on image classification and applied CNN on the \textit{Ekush} dataset, we can visualize the CNN model through \textbf{gradient-weighted class activation maps} or \textbf{Grad-CAM}. Grad-Cam basically works with class activation maps. As the performance of machine learning and neural network-based models are increasing the speed of different complex computations, regression, and classification. But this comes with a huge risk. ~\cite{gehrmann2019visual} We often don't try to know what happens inside the models, as the models are like black boxes to us and we, the users, are forced to trust the models. We also depend and rely on the models without any hesitation. It would be great if we could see on what basis a model generates any type of prediction. \\
\begin{figure}
		\centering
		\begin{subfigure}[b]{.31\columnwidth}
		    \centering
			\includegraphics[width=1\linewidth]{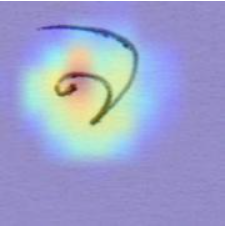}
		\end{subfigure}
		\begin{subfigure}[b]{.31\columnwidth}
		    \centering
			\includegraphics[width=1\linewidth]{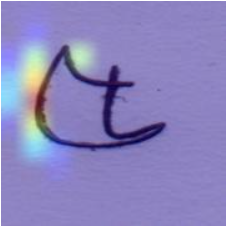}
		\end{subfigure}
		\begin{subfigure}[b]{.31\columnwidth}
		    \centering
			\includegraphics[width=1\linewidth]{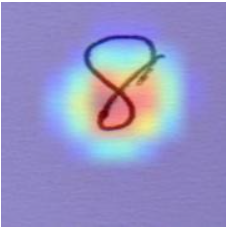}
		\end{subfigure}
		
		\setlength{\belowcaptionskip}{-14pt}
		\caption{Grad-CAM output on 3 sample of \textit{NumtaDB} dataset using the pretrained model on \textbf{imagenet} dataset}
		\label{fig:imagenet}
	\end{figure}
There have been numerous theories proposed to explain the model's behavior. The actual implementation of these approaches is Keras-vis. For visualization, it can be used with Keras models. The Grad-CAM class activation maps, which generate heatmaps at the latent convolutional level instead of the compact layer level, are one of the visualizations included with Keras-vis. More spatial details are taken into account during this whole process ~\cite{cam}.
    \begin{figure*}[ht]
		\centering
			\includegraphics[width=1\linewidth]{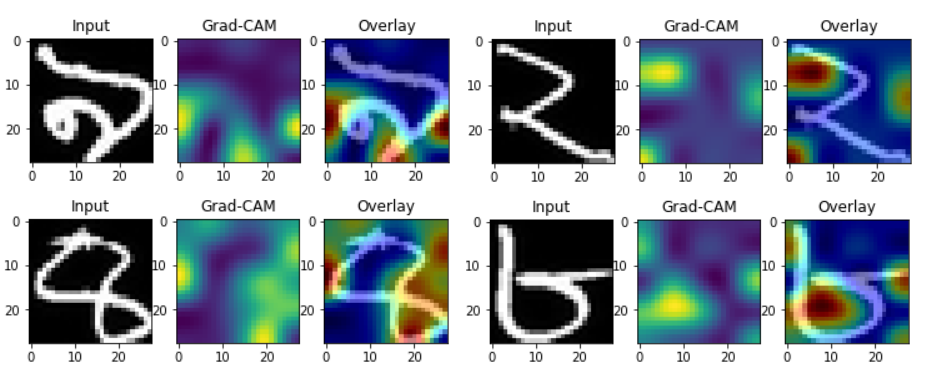}
		\caption{Some output of Grad-CAM on \textit{Ekush} dataset using CNN. The input image is at the \textit{left}, heatmap generated by Grad-CAM is at the \textit{middle} and the superimposed image is at the \textit{right} for each three groups of images}
		\label{fig:grad-CAM}
	\end{figure*}
    Grad-CAM is quite different than traditional CAM. Traditional CAM can be used by small ConvNets which are without dense layers, uninterruptedly passing ahead the convolutional feature maps to the final layer ~\cite{8237336}. But Grad-CAM is based on saliency maps, 
which tells us about the significance of the pixels of a given image. 
In Grad-CAM the gradient of the output layer, which is the class prediction layer, is computed concerning the feature maps of the final layer at first and replaced with the linear function in the implementation. Then the gradients subside and measure the comparative significance of these feature maps for making the class forecasting using the average pooling. A gradient-weighted CAM heatmap depicting positive and negative key elements for the input picture is constructed after producing a linear combination of the feature maps and their ratings. Those are the areas that likely contain the region of interest. Finally, the heatmap was run through a \textbf{ReLU} function to remove the negative areas, setting their relevance to zero, and maintaining just the positive areas' relevance. ~\cite{8237336} ~\cite{cam}. \\	

    We first applied Grad-CAM on some of the data of \textit{NumtaDB} dataset using the weights of a pretrained model on the \textbf{imagenet} dataset \cite{5206848}. Here are some outputs in Fig ~\ref{fig:imagenet}. The output was quite satisfactory, as we can see in these superimposed images.

  After applying Grad-CAM in our CNN model fitted on \textit{Ekush} dataset, we generated some heatmaps on some of our train images randomly and got a superimposed image from the overlay function using the grad cam heatmap and the blue channel of the images. We have shown some of the results here in Fig ~\ref{fig:grad-CAM}. We can see here in the figure that the area of interest from where the CNN model makes the prediction is colorized in the heatmap, and we also tried to show whether the heatmap is generating correctly or not in the superimposed image. We can now visualize the inner perspective of the model using this Grad-CAM XAI method for \textit{Ekush} dataset.

	\section{Limitations And Future Work}
    There are a lot of datasets on Bengali handwritten characters. Due to time constraints, we have just chosen 2 datasets and used only the Bengali digit. Besides, the XAI method we used, which is Grad-Cam, only works on gradient-based models like neural networks. That is why we could not apply this XAI method to different machine learning methods like decision trees or random forests, etc. We can explore some new dimensions in the future, like (i) finding a more accurate model for the used dataset (ii) working with other datasets created on Bengali handwritten digits, and (iii) taking the characters into consideration and applying different algorithms on them.

	\section{Conclusion}
    In our work, we tried to find out some decent accuracy for the \textit{Ekush} and \textit{NumtaDB} datasets. For the second dataset, the accuracy is not so good, but we showed some models like NuSVC and K Nearest Neighbour give us a decent accuracy. Applying Grad-CAM to this dataset was a good idea. It gives us a great visualization of the region of interest, although the weights are from a pretrained model. The \textit{Ekush} dataset gave us some good results. We got a great accuracy of $91.50\%$ using the Random Forest classifier. Besides, CNN has given us $96.71\%$ testing accuracy from the \textit{Ekush} dataset. We were also able to see some great visualization of our CNN model on some randomly selected images from the test dataset using the Grad-CAM XAI method. Working with complex Bengali characters could be great future work. We believe that this study will add some value in the field of handwritten character recognition.
	
	
	\bibliographystyle{unsrt}
	\bibliography{ref}

\end{document}